\documentclass[letterpaper, 10 pt, conference]{ieeeconf}  

\IEEEoverridecommandlockouts                              

\usepackage{times}
\usepackage{epsfig}
\usepackage{graphicx}
\usepackage{amsmath}
\usepackage{amssymb}
\usepackage{algorithm}
\usepackage{algorithmic}
\usepackage{mathrsfs}
\usepackage{multirow}

\usepackage{amsthm}
\theoremstyle{plain}

\usepackage[section]{placeins}

\usepackage[]{animate}

\usepackage{graphicx}
\usepackage{animate}

\usepackage{float}

\usepackage{multirow}
\usepackage{rotating}
\usepackage{wrapfig}
\usepackage{lineno}
\usepackage{epsfig}
\usepackage{amsmath,amssymb}
\usepackage{color}
\usepackage{booktabs}       
\usepackage{amsfonts,amsmath}       
\usepackage{nicefrac}       
\usepackage{microtype} 
\usepackage{float}
\usepackage{amsmath}
 
\usepackage{enumerate}

\title{\LARGE \bf
Deep 3D-CNN for Depression Diagnosis with Facial Video Recording of Self-Rating Depression Scale Questionnaire
}

\author{ Wanqing Xie$^{\dag}$, Lizhong Liang$^{\dag}$, Yao Lu, Hui Luo$^{*}$, Xiaofeng Liu$^{*}$
\thanks{$^{\dag}$ W. Xie and L. Liang contribute equally.}
\thanks{$^{*}$ H. Luo and X. Liu are the corresponding authors.}
\thanks{Wanqing Xie is with the Harbin Engineering University and Suzhou Fanhan Information Technology Company.}
\thanks{L. Liang  is with the Sun Yat-sen University.}
\thanks{Yao Lu is with the Sun Yat-sen University.}
\thanks{Hui Luo is with the Southern Marine Science and Engineering Guangdong Laboratory (Zhanjiang) and the Marine Biomedical Research Institute of Guangdong.}
\thanks{X. Liu is with the Harvard Medical School and Suzhou Fanhan Information Technology Company (xliu11@bidmc.harvard.edu)}

}

\begin{document}

\maketitle
\thispagestyle{empty}
\pagestyle{empty}

\begin{abstract}
The Self-Rating Depression Scale (SDS) questionnaire is commonly utilized for effective depression preliminary screening. The uncontrolled self-administered measure, on the other hand, maybe readily influenced by insouciant or dishonest responses, yielding different findings from the clinician-administered diagnostic. Facial expression (FE) and behaviors are important in clinician-administered assessments, but they are underappreciated in self-administered evaluations. We use a new dataset of 200 participants to demonstrate the validity of self-rating questionnaires and their accompanying question-by-question video recordings in this study. We offer an end-to-end system to handle the face video recording that is conditioned on the questionnaire answers and the responding time to automatically interpret sadness from the SDS assessment and the associated video. We modified a 3D-CNN for temporal feature extraction and compared various state-of-the-art temporal modeling techniques. The superior performance of our system shows the validity of combining facial video recording with the SDS score for more accurate self-diagnose.
\newline
 
\end{abstract}

\section{Introduction}

\begin{figure}[t]
\begin{center}\vspace{+5pt}
\includegraphics[width=1\linewidth]{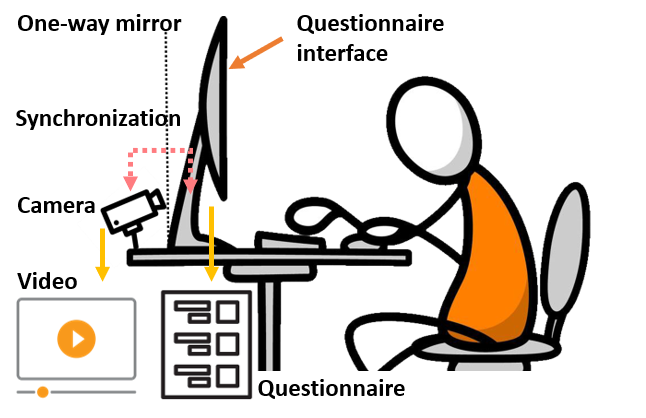} 
\end{center} 
\caption{The graphic depicts the process of collecting Self-Rating Depression Scale questionnaires and the synchronizing facial recordings that accompany them. In order to pick up on the depressive symptoms, we may use SDS score as well as facial expression films.}
\label{fig1}
\end{figure}

While depression is a widespread and severe mental health condition, it is treatable \cite{beck2009depression}. Early accurate detection may help with intervention treatment. The sooner you can start therapy, the more successful it will be \cite{beck2009depression}. However, the problem is that the complete clinical interview for final diagnose may be expensive for large-scale screening. \cite{xie2021interpreting,paykel1985clinical}.

The Self-Rating Depression Scale (SDS) \cite{zung1965self1} is a user-administrated, 20-question screen questionnaires that has gained widespread acceptance in the medical community. The SDS result, on the other hand, may differ from the clinical interviews used for confirming the depression diagnose \cite{gabrys1985reliability}. One explanation is that the uncontrolled self-administered assessment can be readily influenced by indolent or dishonest responses \cite{zung1965self1}, yielding different findings than the clinician-administered interview.

When it comes to clinical assessment, facial expressions (FE) \cite{liu2017adaptive,liu2019hard} and body-actions \cite{pampouchidou2017automatic} may take a significant position. However, when it comes to self-administered evaluation, FE and actions have received less attention. In fact, for many psychiatric studies, facial expressions and behaviors may serve as expressive characteristics \cite{krause1989facial}. Based on this understanding, we gather a new dataset of 200 participants to demonstrate the effectiveness of self rating questionnaire with videos recording in this work. The video is recorded using the software defined camera (SDC) system that is synced with a questionnaires application, beginning with the question display and terminating when a score is chosen. This allows for a more precise connection between the survey and the video. Furthermore, the length of time it takes to react may influence the diagnose \cite{lewinsohn1969depression}. 

With the fast development of deep learning for recognition \cite{liu2021mutual,liu2019unimodal,liu2020unimodal,he2020classification,liu2018ordinal,liu2019conservative,liu2019feature,che2019deep}, a hierarchical approach is developed for automatically interpreting depression based on the SDS assessment, its associated FE, and action video recording, among other things. To be more specific, we effectively extract the temporal information from each question-wise video by adjusting the 3D convolutional neural networks to the particular question (3D-CNN) \cite{liu2019dependency}. Aside from that, we conducted a thorough examination of widely recognized models such as the recursive network (RNN) \cite{8121994}, and the non-local network \cite{liu2018dependency,liu2019permutation}. After that, a question-by-question video feature is combined with the questionnaire scores to arrive at a final depression result.

The validity of combining SDS with video recording, as well as the performance of our temporal modeling, are further shown by our comprehensive examinations.

\begin{figure}[t]
\begin{center}
\includegraphics[width=0.9\linewidth]{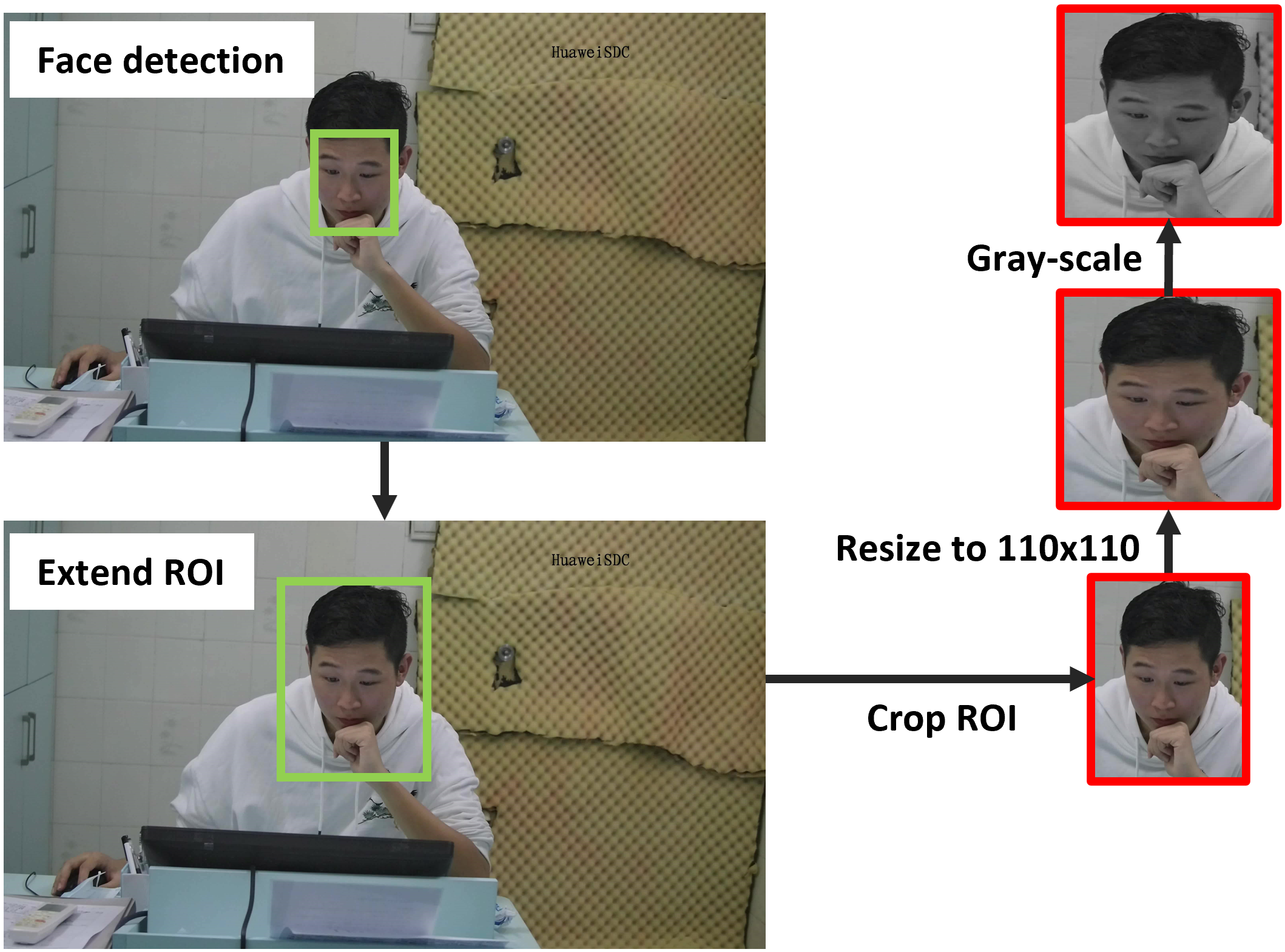} 
\end{center} 
\caption{The collection of SDS questions and the synchronized facial films that accompany them. Both the SDS score and facial expression films may aid in the detection of depression.}
\label{figexp2}
\end{figure}

\section{Methodology}

\begin{table}[t]  \caption{The statistics of the 200 subjects who were interviewed.}\label{tab1}
\centering
\resizebox{0.7\linewidth}{!}{
\begin{tabular}{l|l|c}
\hline\hline
Final diagnose&	SDS result & Number\\\hline
Normal	& Normal&	86\\ 
Normal	&Depression &20\\ 
Depression &Normal	&20\\ 
Depression&Depression&	74\\\hline\hline
\end{tabular}} 
\end{table}

\subsection{Data collection and its processing}

We compile the SDS questionnaires filled by 200 people and their corresponding face video shot with an HD camera. It was authorized by the Guangdong Medical University Affiliated Hospital's Ethics Committee (No. PJ2021-026).

To prevent being influenced by others, participants are advised to stay in a closed consultation office and complete these questionnaires according to the instructions shown in the computer screen. Additionally, a software defined camera (SDC) is concealed behind a one-way mirror, and participants are unaware of the camera's existence throughout the test. The program connects the camera to the questionnaire, and syncs the camera up to begin recording when the question used to start the video is shown, and then stop the video when the decision is made. The question-wise video collection protocol is illustrated in Fig. \ref{fig1}.

In order to make sure that our study area only includes the participants' faces, we utilize the face detector to cut off the facial region \cite{zhang2017faceboxes}. We also increase the face region by about 120\% and enlarge the expanded rectangle of interests of each image to 110$\times$110 to include hand actions such as head scratching \cite{carpenter2000psychotic} and chin touching \cite{kazdin1985nonverbal}. Additionally, gray processing is used to decrease the input channel size. The flowchart that precedes the pre-processing is shown in Fig. \ref{figexp2}.

Followed by the the self administering evaluations, the clinical interviews are performed to ensure the correctness of the diagnose of depression \cite{paykel1985clinical}. In Table \ref{tab1}, we provide an extensive set of statistical statistics about the SDS, as well as our 200 participants' final results.

\subsection{Hierarchical conditional framework}
There is much emotional information included in long-term video recordings of SDS evaluations \cite{liu2018adaptive,liu2021identity,liu2021mutual}, but these recordings present a number of processing difficulties. Because it is redundant \cite{liu2018adaptive,liu2021identity,liu2021mutual}, and only a limited number of (i.e., sparse) frames are helpful for depression detection, it is recommended that a threshold of four or more frames be used. In this work, we uniformly sampled 100 frames in each question-wise video. We denote the $i-th$ frame in the $q-th$ question as $I^q_i$. Other than the RNNs \cite{8121994} and the non-local network \cite{liu2018dependency,liu2019permutation}, the 3D-CNN module may be a better tool for efficient processing. \cite{liu2019dependency}.

The response time is added to the answers of the question-wise video features $\{a^q\}_{q=1}^{20}$, which are then concatenated with the matching tabular questionnaire findings. For the SDS, we can use four-dimensional one-hot vectors to represent the kind of issue since we notice that the SDS has four scales. When the responding time is added to the equation, the result is likewise a scalar, only one dimension longer. As for binary classification, we utilize the fully linked layers with the sigmoid output unit for normal or depression.

We notice that all of the videos and questions use the same 3D-CNN or self attention module. We describe the structure for our hierarchically built 3D-CNN and question-level fusion module in the following subsections.

\begin{figure}[t]
\begin{center}
\includegraphics[width=0.9\linewidth]{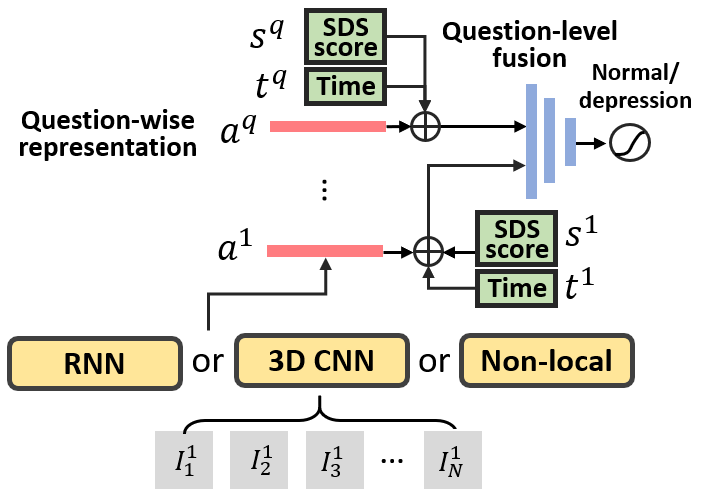} 
\end{center} 
\caption{The detailed structure of the hierarchical conditional framework.}
\label{fig2}
\end{figure}

\subsection{3D-CNN for temporal exploration}

\begin{figure}[t]
\begin{center}
\includegraphics[width=0.9\linewidth]{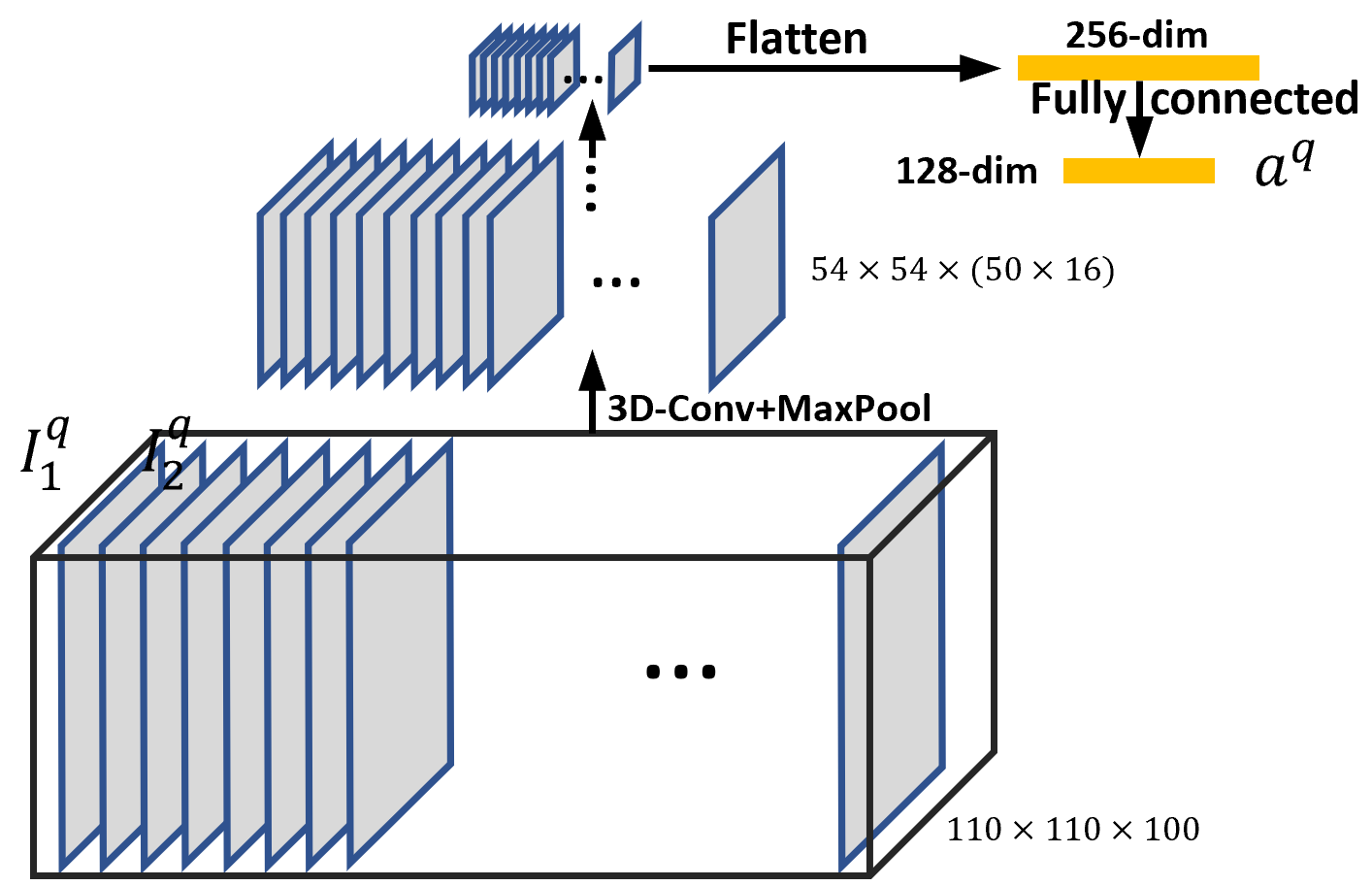}  
\end{center} 
\caption{The 3D-CNN model for temporal feature exploration.}
\label{fig3}
\end{figure}

3D-CNNs have been shown to be successful in extracting rapid temporal representations from relatively brief fixed-length video \cite{ji20123d}. We describe the connections between consecutive frames using the conventional 3D convolution technique. In Fig. \ref{fig3}, we depict the fundamental structure of 3D-CNN. Following a couple 3D CNNs and max-pooling layers, we pass the 256 dimension vector to a FC layer, resulting in the 128 dimension representation $a^q$ for each clip, where $q$ indexes the twenty questions.
 
The max-pooling layers reduce the height, breadth, and frame dimensions to half their original values and are called (2$\times$2$\times$2)-(1$\times$1$\times$1). Table \ref{tab2} depicts the network structure.

\subsection{Question wise conditional fusion}   

Each question's SDS questionnaire score is represented by $s^q\in\mathbb{R}^4$. Additionally, we discovered experimentally that the time spent on answering each question might be beneficial for diagnose. As a result, the duration of each video question $t^q\in\mathbb{R}$ has been concatenated. For a three-second video, we set $t^q$ to three. A response that is either too brief or too lengthy may be untrustworthy \cite{lewinsohn1969depression}.

Then, $a_q$, $s_q$, $t_q$, and $t_q$ are concatenated for creating a 133 dimensional representation for each of the questions.

In order to arrive at a final depression diagnose, the feature consisting the attributes of 20 question-by-question videos and questionnaires is concatenated with a 2660 dimensional representation. The completely linked layers are used, and Table \ref{tab3} has the detailed network topology.
Between each FC layer, the ReLUs are utilized as our non-linear mapping functions.
 
For binary classification, we use the sigmoid function for final prediction $p=\frac{1}{1+e^{out}}\in(0,1)$, where $out$ is the prediction's scalar of the  final layer in the networks, which is normalized to the probability that this subject is a depression sufferer. We highlight that we indicate healthy subject with a value of 0 (i.e., $y=0$) and depressed subject with a value of 1 (i.e., $y=1$) based on the clinician-based interview. We utilize the cross entropy loss for optimizing and train our model automatically through backpropagation.
\begin{align}
   \mathcal{L}= -y\text{log}(p)-(1-y)\text{log}(1-p),
\end{align}
which incurs no punishment if $p$ is equal to its equivalent $y$. Additionally, we set the binary classification threshold as $p=0.5$ in testing.

\section{Experiments}

In experiments, we compare the classification performance of ours 3D-CNN model to that of RNNs and non-local models. Additionally, we offer systematic ablation research and sensitivity analysis to show the efficacy of our framework's design decision.

\subsection{Implementation Details and metrics}

All tests were conducted on our server, which was equipped with an NVIDIA V100 GPU, a Xeon E5 v4 CPU, and 128GB RAM. Our model is trained  with the Adam \cite{kingma2014adam} optimizer with the hyper-parameters of $\beta_{1}$=0.9 and $\beta_{2}$=0.999. For our dataset, we chose a batch size of two. Our framework's and comparable techniques' networks are trained for 200 epochs to provide a fair comparison. The results of five random initializations are shown, together with the standard deviation and average performance. We point out that the 3D-CNN and the redundancy-aware self-attention modules are shared across all of the queries, allowing them to be processed in parallel. Subjects' average inference time is just 1.3 seconds, while the training takes about 8 hours.

\begin{table}[t]
\caption{Our 3D-CNN module's comprehensive structure for temporal feature extraction.}  
\centering 
\resizebox{1\columnwidth}{!}{%
\begin{tabular}{l | l | l} 
\hline\hline 
Input Size&Type& Filter Shape   \\ [0.5ex] 
\hline 

$[110\times110\times100]$&3D-Conv& [16 kernels of $3\times3\times3$]\\
$[108\times108\times(100\times16)]$&MaxPool& (2$\times$2$\times$2)-(1$\times$1$\times$1)\\
\hline

$[54\times54\times(50\times16)]$&3D-Conv& [32 kernels of $3\times3\times3$]\\
$[52\times52\times(50\times32)]$&MaxPool& (2$\times$2$\times$2)-(1$\times$1$\times$1)\\
\hline

$[26\times26\times(25\times32)]$& 3D-Conv& [64 kernels of $3\times3\times3$]\\ 
$[24\times24\times(25\times64)]$&MaxPool& (2$\times$2$\times$2)-(1$\times$1$\times$1)\\
\hline

$[12\times12\times(13\times64)]$& 3D-Conv& [128 kernels of $3\times3\times3$]\\
$[10\times10\times(13\times128)]$&MaxPool& (2$\times$2$\times$2)-(1$\times$1$\times$1)\\
\hline

$[5\times5\times(6\times128)]$& 3D-Conv& [256 kernels of $5\times5\times6$]\\

\hline

$[1\times1\times(1\times256)]$ & Flatten & N/A\\\hline

256-dim &  FC & 128-dim\\

\hline\hline
 
\end{tabular}
\label{tab2} 
} 
\end{table}

For the 200 individuals in our dataset, we use fivefold cross-validation. We divided the dataset into five subgroups, each with 40 individuals. We notice that there is no overlap between the two folds in terms of topics. Then, we choose a fold sequentially as our testing set (i.e., 40 people), while using the other four folds (i.e., 160 participants) for training.

\begin{table}[t]
\caption{Our question-level fusion's comprehensive, fully-connected structure.} 
\centering 
\resizebox{0.8\columnwidth}{!}{%
\begin{tabular}{l | l | l} 
\hline\hline 
Input Size&Type& Filter Shape   \\ [0.5ex] 
\hline 

[128+4+1]$\times$20 &Concatenate& 2660 \\
2660 &FC& 1024 with ReLU \\
1024 &FC& 256 with ReLU \\
256 &FC& 1 with sigmoid \\

\hline\hline
 
\end{tabular}
\label{tab3} 
} 
\end{table}

\subsection{Baselines and Comparison results}

When choosing a modality for the video and sound dataset, there are three choices: (1) the video alone, (2) the sound alone, or (3) use the two together. With only the SDS assessment results, we can simply combine the scores from the 20 questions and use the 50-point threshold for healthy and depression categorization. As shown in Table \ref{tab1}, SDS findings may vary from physician diagnose. Additionally, we attempted to do classification only using the video modality, which did not concatenate $q$ for the fusion of question levels.

\begin{table}[t]
\caption{The performance of binary classification is compared.}  
\centering 
\resizebox{1\columnwidth}{!}{%
\begin{tabular}{l | c | c | c} 
\hline\hline 
Methods&Accuracy& Sensitivity & Specificity \\ [0.5ex] 
\hline 
$[\text{SDS only}]$sum & 0.800$\pm$0.000 & 0.787$\pm$0.000 & 0.811$\pm$0.000\\

$[\text{Video only}]$3D-CNN &  0.664$\pm$0.013   & 0.657$\pm$0.011   &  0.708$\pm$0.009  \\

$[\text{SDS+Video}]$RNN &  0.840$\pm$0.007  &   0.816$\pm$0.007 &   0.846$\pm$0.006  \\
$[\text{SDS+Video}]$non-local &   0.885$\pm$0.005  &  0.859$\pm$0.003  &  0.893$\pm$0.007   \\

$[\text{SDS+Video}]$3D-CNN &  \textbf{0.903$\pm$0.006}   & \textbf{0.895$\pm$0.005}   &  \textbf{0.908$\pm$0.007}  \\
\hline\hline
\end{tabular}
\label{resulttab2} 
} 
\end{table}   

It is evident that combining SDS and video can significantly exceed SDS only, demonstrating the efficacy of the extra video modality. The accompanying video provides additional information that aids in the correct detection of depression. Additionally, we are able to forecast sadness with an accuracy of 0.69 when we just utilize the video, which is greater than the chance likelihood of 0.5.

To illustrate our model's efficacy, we compared it to two baseline approaches, i.e., RNN \cite{8121994} and non-local networks \cite{liu2018dependency,liu2019permutation}. Notably, this is the first time that RNNs and non-local networks have been used for depression diagnose.

In this instance, RNNs are the traditional option for temporal data exploration \cite{8121994}. We have developed a bi-directional LSTM network \cite{8121994} specifically for the purpose of video feature extraction. The non-local strategy \cite{liu2018dependency,liu2019permutation} is currently in development to help tackle the long-term memory loss that is seen in RNN \cite{pascanu2013difficulty}. To extract the 128 dimensional question-wise feature representation, we utilize an RNNs or non-local as the replacement for 3D-CNN. See Table \ref{resulttab2} for the quantitative assessment findings. We obtain better performance by using our developed framework rather than the RNN or non-locals.

\section{Conclusions}

This paper target to automate the Self-Rating Depression Scale evaluation and question-by-question video recording exploration. We expand the face detection box considers facial expression, eye movement, head-scratching, and chin-touching motions. The hierarchical neural networks architecture is built to evaluate the question-by-question video. As a result, our method is very accurate in identifying depression, especially on smartphones. Positive individuals will be sent to a specialized facility for medical diagnose and treatment.

\section*{Acknowledgment}
This work was partially supported by NIH [NS061841], Jiangsu NSF [BK20200238], and Southern Marine Science and Engineering Guangdong Laboratory [ZJW-2019-007].




%

\bibliographystyle{IEEEtran}
\bibliography{main}

\end{document}